\documentclass[letterpaper, 10 pt, conference]{ieeeconf} 

\IEEEoverridecommandlockouts                        

\usepackage{graphics}
\usepackage{amsmath} 
\usepackage{amssymb}
\usepackage{cite}
\usepackage{tikz}
\usepackage{amsmath}
\usepackage{tabularx}
\usepackage{array}
\usepackage{booktabs}
\usepackage{bm}
\usepackage{algorithm}
\usepackage{algorithmic}
\usepackage{bbm}
\usepackage{adjustbox}
\usepackage{caption}

\usepackage{tablefootnote}

\usepackage{hyperref}

\definecolor{green}{RGB}{11,155,13}

\DeclareMathOperator*{\argmin}{argmin}

\title{\LARGE \bf
Toward Wheeled Mobility on Vertically Challenging Terrain:\\Platforms, Datasets, and Algorithms
}

\author{Aniket Datar$^*$, Chenhui Pan$^*$, Mohammad Nazeri, and Xuesu Xiao
\thanks{$^*$Equally contributing authors. All authors are with the Department of Computer Science, George Mason University {\tt\scriptsize \{adatar, cpan7, mnazerir, xiao\}@gmu.edu}}
}

\begin{document}
\maketitle
\thispagestyle{empty}
\pagestyle{empty}

\begin{abstract}
Most conventional wheeled robots can only move in flat environments and simply divide their planar workspaces into free spaces and obstacles. Deeming obstacles as non-traversable significantly limits wheeled robots' mobility in real-world, extremely rugged, off-road environments, where part of the terrain (e.g., irregular boulders and fallen trees) will be treated as non-traversable obstacles. To improve wheeled mobility in those environments with \emph{vertically challenging terrain}, we present two wheeled platforms with little hardware modification compared to conventional wheeled robots; we collect datasets of our wheeled robots crawling over previously non-traversable, vertically challenging terrain to facilitate data-driven mobility; we also present algorithms and their experimental results to show that conventional wheeled robots have previously unrealized potential of moving through vertically challenging terrain. We make our platforms, datasets, and algorithms publicly available to facilitate future research on wheeled mobility.\footnote{\scriptsize Website: \url{https://cs.gmu.edu/~xiao/Research/Verti-Wheelers/}}
\end{abstract}

\section{INTRODUCTION}
\label{sec::introduction}
Building mobile robots that are capable of reaching as many places as possible has long been a dream for many robotics researchers. Indeed, autonomous mobile robots have ventured into remote deserts for scientific exploration~\cite{wettergreen2010science}, explored extraterrestrial planets to look for signs of life~\cite{maurette2003mars, islam2017novel}, and assisted with search and rescue missions in hazardous or difficult-to-reach environments~\cite{xiao2015locomotive, murphy2016two, xiao2017uav, xiao2021autonomous}. 

One particular thrust in this area of research is the development of ground robots capable of navigating vertically challenging terrain (e.g., steep slopes, rocky outcroppings, and uneven surfaces)~\cite{murphy2014disaster, xiao2018review, mcgarey2016system}. Achieving reliable and robust mobility in these environments is challenging due to the intricate nature of the terrain, the complex vehicle-terrain interactions, the adverse impact caused by gravity, the potential deformation of the vehicle chassis, and the varing traction between the wheels and the terrain. Despite these difficulties, such mobility has been made possible mainly through advancement in hardware, including the development of specialized robot platforms~\cite{xiao2015locomotive, mcgarey2016system, zheng2022mathbf, islam2017novel, xu2023whole} and the use of new materials ~\cite{liu2018anyclimb}. These robots are capable of climbing walls~\cite{liu2018anyclimb}, scaling cliffs~\cite{mcgarey2016system}, and traversing rough terrain with ease~\cite{zheng2022mathbf, xu2023whole}, making them suitable for a wide range of real-world applications.

However, despite these hardware advances, the vast majority of currently available ground robots are still conventional wheeled platforms, whose mobility is mostly limited to flat surfaces. Most existing autonomous navigation systems for wheeled robots merely divide their presumably planar workspaces into free spaces (traversable) or obstacles (non-traversable), significantly reducing these robots' reachability in the real world, especially outdoor off-road environments where vertical protrusions from the ground are not uncommon. Our work is motivated by such limitations and aims at expanding the mobility of these widely available wheeled robot platforms so that they can venture into vertically challenging environments, which would otherwise be deemed as obstacles (non-traversable) or require specialized hardware. Note that the \emph{vertically challenging} terrain we are interested in conquering requires driving wheeled robots over irregular and complex obstacles and is therefore much more difficult compared to simply driving on \emph{non-flat} environments.

\begin{figure}[t]
  \centering 
  \includegraphics[width=\columnwidth]{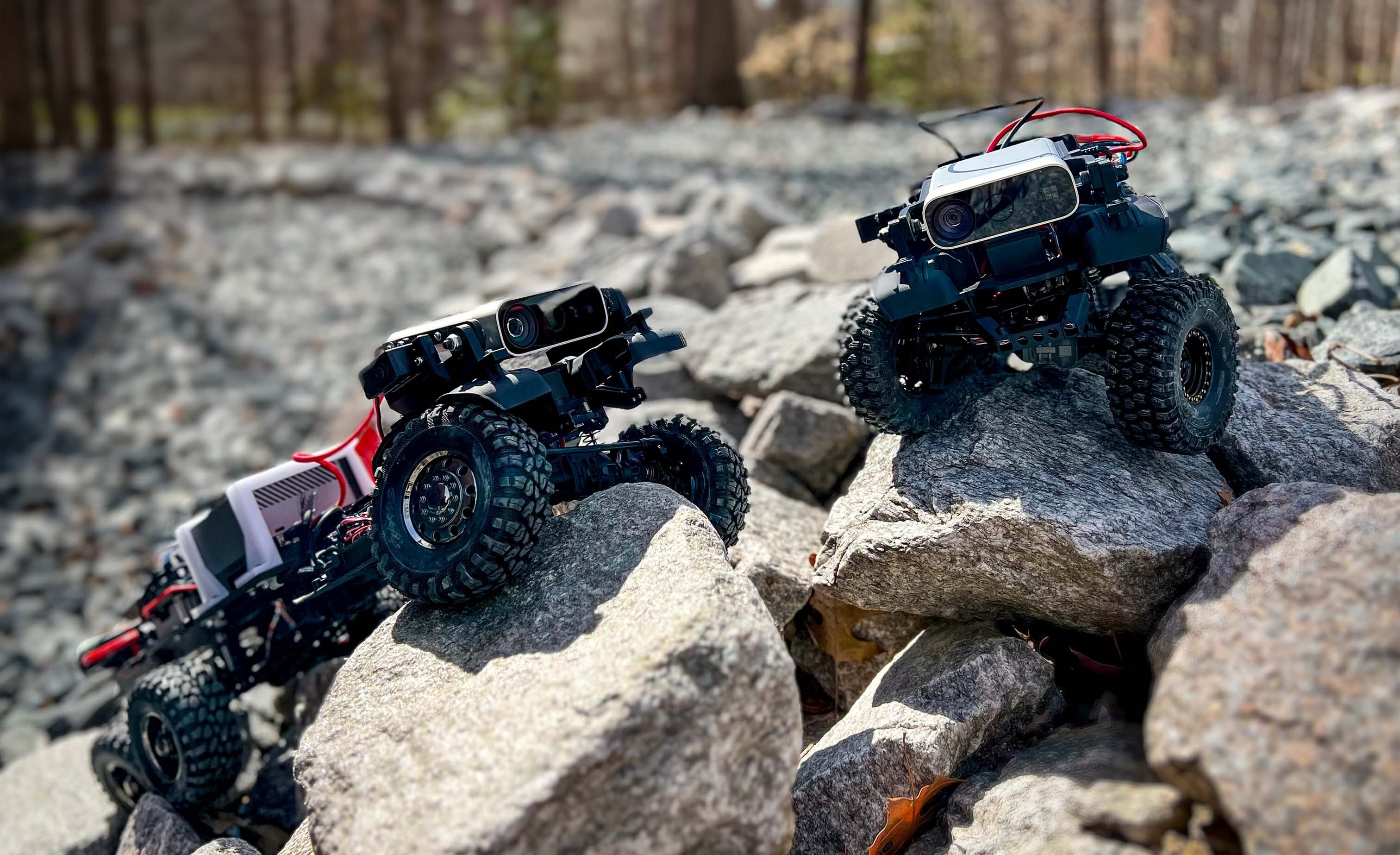}
  \caption{The Verti-Wheelers: Conventional Wheeled Vehicles Moving through Vertically Challenging Terrain. }
  \label{fig::VW}
\end{figure}

To this end, we first present an open-source design of two wheeled robot platforms, the Verti-Wheelers, which are representative of the majority of existing conventional ground mobile robot platforms, and hypothesize that conventional wheeled robots can also navigate many vertically challenging terrain, which are normally considered as non-traversable obstacles (Fig.~\ref{fig::VW}). Second, considering the difficulty in creating accurate analytical models to solve this problem, we collect and make publicly available datasets to facilitate data-driven approaches. Third, we present three algorithms to autonomously drive wheeled robots over vertically challenging terrain: an open-loop, a classical rule-based, and an end-to-end learning-based approach, which can be used as baselines to benchmark future research. We provide extensive experimental results of all three algorithms in both an indoor controlled testbed and outdoor natural environments, confirm our hypothesis, and point out future research directions for wheeled mobility on vertically challenging terrain. To the best of our knowledge, no existing work has tackled vertically challenging terrain with conventional wheeled robots in the real world.

\section{RELATED WORK}
\label{sec::related_work}
This section discusses conventional wheeled robot mobility, reviews novel hardware to expand ground mobility beyond the capability of conventional wheeled platforms, and surveys data-driven approaches to improve robot mobility. 

\subsection{Conventional Wheeled Mobility}
Due to their simplicity and efficiency, wheeled robots have been widely used in various applications such as scientific exploration~\cite{wettergreen2010science}, autonomous delivery~\cite{starship}, and search and rescue~\cite{murphy2014disaster}.
Equipped with differential-drive mechanism~\cite{malu2014kinematics}, Ackermann steering~\cite{atreya2022high}, or omnidirectional wheels~\cite{lu2022design}, these robots can move through their planar workspaces and reach their goals. 

One basic capability for wheeled robots is obstacle avoidance. Roboticists have developed autonomous navigation systems which first divide their planar workspaces into free spaces and obstacles and then move robots to their goals without collisions~\cite{fox1997dynamic, quinlan1993elastic, xiao2022autonomous, perille2020benchmarking, nair2022dynabarn}. Researchers have also investigated off-road navigation~\cite{pan2020imitation, xiao2021learning, sivaprakasam2021improving, karnan2022vi}, in which the planar workspace includes a variety of terrain types, such as gravel, grass, mud, sand, and snow. Instead of the binary free/obstacle depiction of the planner workspace, off-road navigation systems usually build semantic or traversability maps~\cite{maturana2018real, shaban2022semantic}, e.g., gravel is better to drive on than grass, which is better than mud. However, for vertical protrusions from the ground, e.g., large boulders or fallen tree trunks, these systems still treat them as non-traversable obstacles due to mobility limitations caused by the wheels. 

However, in real-world unstructured environments, vertical protrusions from the ground are not necessarily non-traversable, even for wheeled robots. Consider the motor-sport of rock crawling, in which human drivers are able to drive off-road vehicles over obstacles of similar size to the vehicles themselves. Deeming all these vertically challenging terrain as completely non-traversable largely limits the mobility of wheeled robots in the real world. In this work, we aim at expanding wheeled mobility to these previously impossible, vertically challenging terrain.

\subsection{Novel Hardware for Vertically Challenging Terrain}
One way to improve ground mobility is through novel hardware design. To overcome the limitations of wheels, researchers have developed robots with alternative actuation: wheels can be replaced by legs when facing extremely rugged terrain~\cite{kutzer2010design, bu2022development}; active suspensions are widely used on planetary rovers to achieve better maneuverability, reconfigurability, and therefore mobility by allowing the chassis to actively conform to different underlying terrain~\cite{cordes2014active, jiang2019lateral}; tracked robots are not particularly novel, but the increased track surface contact compared to wheels makes them more appropriate on vertically challenging terrain, while compromising efficiency on flat surfaces. Another thrust is to develop new adhesive materials so that robots can adhere to steep or even vertical surfaces~\cite{liu2018anyclimb}. 

Despite their superior mobility on vertically challenging environments, these costly, specialized hardware require extra engineering effort and may adversely affect vehicles' mobility and efficiency on flat environments. Considering that most ground robots are wheeled with no or passive suspension systems, our research aims at equipping those conventional wheeled platforms with enhanced mobility to autonomously move through vertically challenging terrain.

\subsection{Data-Driven Robot Mobility}
Thanks to the recent advancement in machine learning, data-driven approaches have been leveraged to improve robot mobility~\cite{xiao2022motion}. Researchers have investigated using learning to achieve adaptive navigation in a variety of environments~\cite{xiao2022learning, xiao2022appl, xiao2020appld, wang2021appli, wang2021apple, xu2021applr, liu2021lifelong}, agile navigation in highly constrained spaces~\cite{xiao2021toward, xiao2021agile, wang2021agile}, high-speed off-road navigation~\cite{pan2020imitation, xiao2021learning, sivaprakasam2021improving, karnan2022vi}, visual-only navigation~\cite{karnan2022voila, kahn2020badgr, shah2023vint, stachowicz2023fastrlap}, and socially compliant navigation~\cite{mirsky2021conflict, francis2023principles, karnan2022socially, chen2017socially, xiao2022learning, nguyen2023toward, park2023learning, hart2020using}. Learning from data using either imitation learning~\cite{pfeiffer2017perception, xiao2020appld, karnan2022voila} or reinforcement learning~\cite{xu2023benchmarking, xu2021machine, faust2018prm} removes the necessity of building analytical models of the environments, such as vehicle-terrain or human-robot interactions, and alleviates the burden of crafting delicate cost functions~\cite{xiao2022learning, sikand2022visual} or tuning unintuitive parameters~\cite{xiao2022appl, xiao2020appld, wang2021appli, wang2021apple, xu2021applr}. 

The problem of autonomously driving wheeled robots through vertically challenging terrain shares many aforementioned difficulties. Therefore, we hypothesize that data-driven approaches are one avenue toward enabling enhanced wheeled mobility on previously impossible, vertically challenging terrain. 

\section{PLATFORMS AND DATASETS}
\label{sec::platforms}

\begin{figure}
  \centering
  \includegraphics[width=\columnwidth]{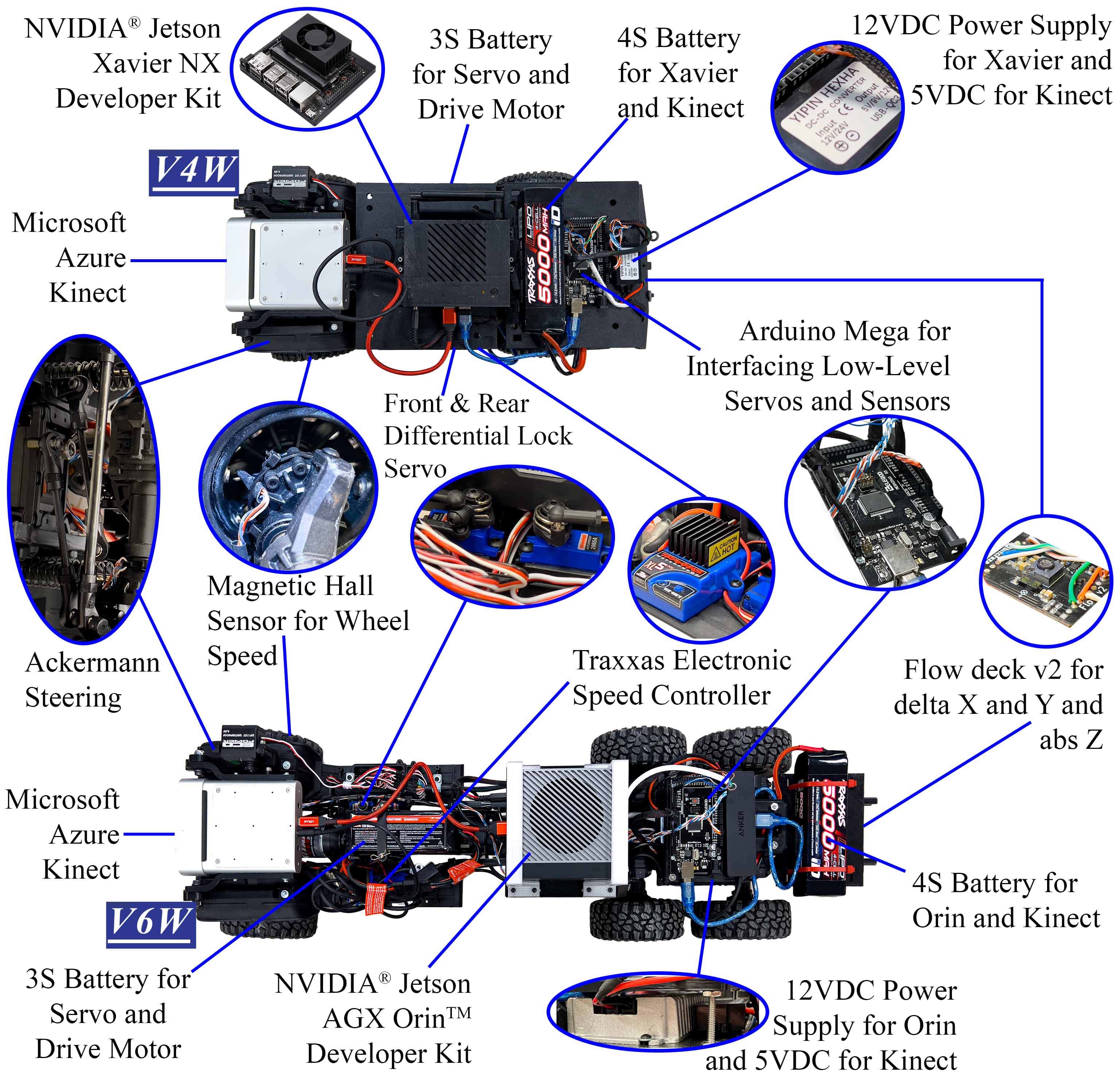}
  \caption{Components of the Verti-Wheelers.}
  \label{fig::layout}
\end{figure}

Considering that most existing mobile robots are wheeled, our goal is to equip these easily available wheeled platforms with no or little specialized hardware modification. To assure it is physically feasible for wheeled platforms to move through vertically challenging terrain, we identify the following seven desiderata for their hardware: 

\noindent \textbf{All-Wheel Drive (D1):} Due to the high possibility of losing wheel contact with the irregular terrain, our wheeled platforms need to be all-wheel drive to maximize traction. 

\noindent \textbf{Independent Suspensions (D2):} Considering the non-flatness of vertically challenging terrain, independent suspensions allow the chassis to conform to and most wheels to contact with the ground to maintain stability and traction. 

\noindent \textbf{Differential Lock (D3):} While traditional differentials aim at increasing efficiency during vehicle turning on flat surfaces, loosing left wheel contact will cause the right wheel to lose traction, or vice versa. Therefore, our wheeled platforms need to be able to lock the differential(s) when necessary.

\noindent \textbf{Low/High Gear (D4):} Although high gear improves energy efficiency on flat surfaces, to mitigate downward force caused by gravity, our platforms require high torque and low gear on vertically challenging terrain. 

\noindent \textbf{Wheel Speed / RPM Sensing (D5):} Precise end-point wheel speed measurement (instead of motor encoder readings that will be skewed by the drivetrain and differential) allows the vehicle to sense wheel slippage before losing wheel contact. 

\noindent \textbf{Ground Speed Sensing (D6):} Our platforms need an easy way to identify when the vehicle is getting stuck, which is very likely to happen on vertically challenging terrain. 

\noindent \textbf{Actuated Perception (D7):} Due to the increased vehicle pose variability (i.e., pitch and roll in addition to yaw) when crawling over vertically challenging terrain, traditional fixed onboard sensors may lose sight of the terrain. Therefore, our platforms need to be able to point their sensors toward the terrain, regardless of how the vehicle pose varies. 

\subsection{Verti-6-Wheeler (V6W)}

For the mechanical components in D1 to D4, we base our platform on an off-the-shelf, three-axle, six-wheel, all-wheel-drive, off-road vehicle chassis from Traxxas. The length of the chassis is 0.863m, with 0.471m front-to-middle and 0.603m front-to-rear axle wheelbase. The total height and width of V6W after outfitting all mechanical and electrical components is 0.200m and 0.249m. D1 and D2 are therefore achieved. We use an Arduino Mega micro-controller to lock/unlock the front and rear differential (D3) and switch between low and high gear (D4) through three servos. For D5, we install four magnetic sensors on the front and middle axles, and eight magnets per wheel to sense the wheel rotation. We choose not to install magnetic sensors on the rear axle considering the rear differential lock is shared by both middle and rear axles. For D6, we install a Crazyflie Flow deck v2 sensor on the chassis facing downward, providing not only 2D ground speed ($x$ and $y$) but also distance between the sensor and the ground ($z$). We choose an Azure Kinect RGB-D camera due to its high-resolution depth perception at close range. For D7, we add a tilt joint for the camera actuated by a servo. We use a complementary filter and camera Inertia Measurement Unit (IMU) readings to estimate the camera orientation and a PID controller to regulate the camera pitch angle. For the core computation unit, we use an NVIDIA Jetson AGX Orin to provide both onboard CPU and GPU computation. To interface all low-level sensors and actuators, we use an Arduino Mega micro-controller.

\begin{figure}
  \centering
  \includegraphics[width=\columnwidth]{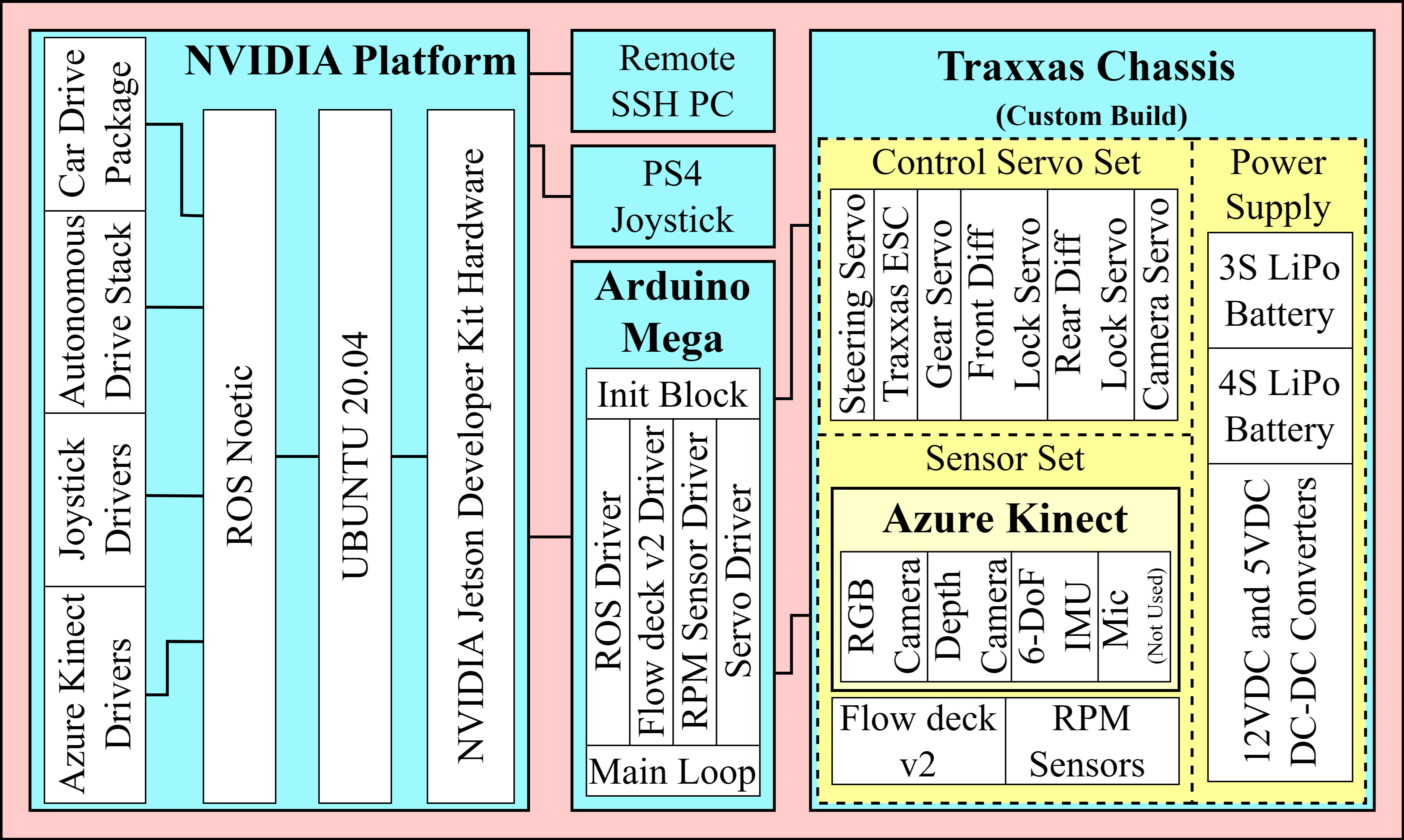}
  \caption{System Architecture of the Verti-Wheelers. }
  \label{fig::architecture}
\end{figure}

\subsection{Verti-4-Wheeler (V4W)}
Considering that most wheeled robots have four wheels, we also build a four-wheeled platform based on an off-the-shelf, two axle, four-wheel-drive, off-road vehicle from Traxxas. The length of the V4W chassis is 0.523m, with a 0.312m wheelbase. The total height and width of V4W is 0.200m and 0.249m. Most components remain the same as the six-wheeler, but to accommodate the small footprint and payload capacity of the four-wheeler, we replace the NVIDIA Jetson AGX Orin with a Xavier NX. The mechanical and electrical components and system architecture for both V4W and V6W are shown in Fig.~\ref{fig::layout} and \ref{fig::architecture} respectively.\footnote{\url{https://github.com/RobotiXX/Verti-Wheelers}}

\subsection{Testbed}
\begin{figure*}
  \centering
  \includegraphics[width=1\textwidth]{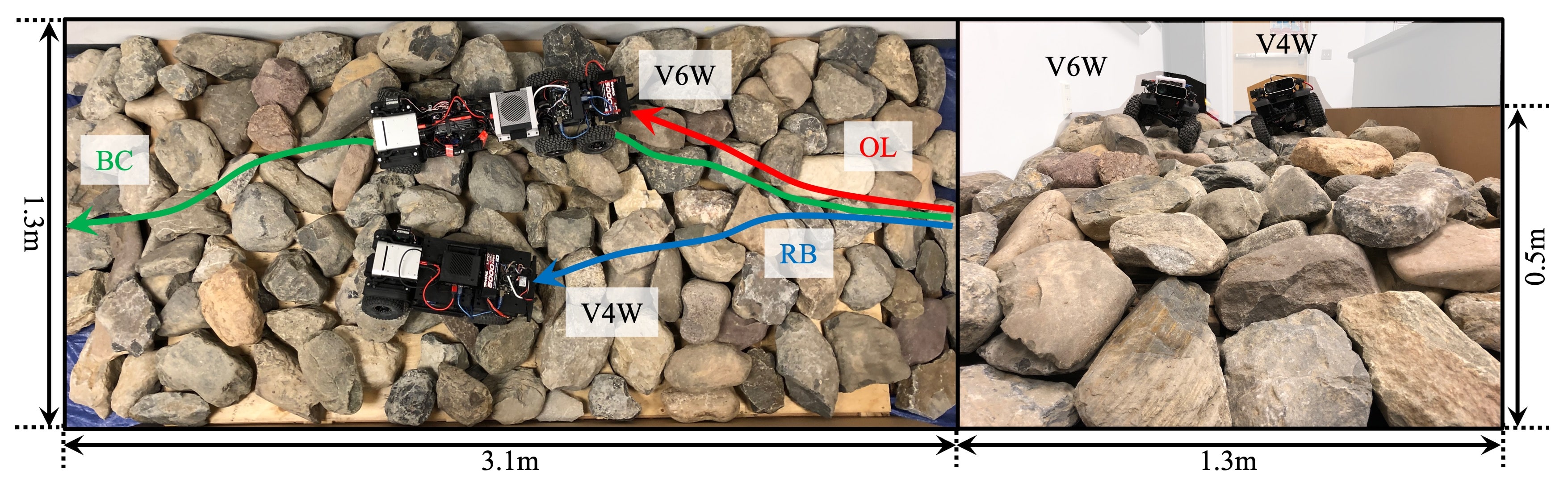}
  \caption{Custom-Built Testbed with V6W and V4W and Example Traversals by the Three Algorithms (OL, RB, and BC).}
  \label{fig::testbed}
\end{figure*}

In addition to outdoor field experiments, we construct a custom-built indoor testbed for vertically challenging terrain for controlled and repeatable experiments: hundreds of rocks and boulders of an average size of 30cm (at the same scale of the V6W and V4W) are randomly laid out and stacked up on a 3.1$\times$1.3m test course. The highest elevation of the test course can reach up to 0.5m, more than twice the height of both vehicles, as shown in Fig.~\ref{fig::testbed}. The testbed is highly reconfigurable by shuffling the pose of each rock/boulder. 

\subsection{Datasets}
Considering the difficulty in representing surface topography and modeling complex vehicle dynamics and the recent success in data-driven mobility~\cite{xiao2022motion}, we collect two datasets with the two wheeled robots on our custom-built testbed.\footnote{\url{https://dataverse.orc.gmu.edu/dataset.xhtml?persistentId=doi:10.13021/orc2020/QSN50Q}} We reconfigure our testbed multiple times and both robots are manually driven through different vertically challenging terrain. We collect the following data streams from the onboard sensors and human teleoperation commands: RGB ($1280\times720\times3$) and depth ($512\times512$) images $i$, wheel speed $w$ (4D float vector for four wheels), ground speed $g$ (relative movement indicators along $\Delta x$ and $\Delta y$ and displacement along $z$, along with two binary reliability indicators for speeds and displacement), differential release/lock $d$ (2D binary vector for both front and rear differentials), low/high gear switch $s$ (1D binary vector), linear velocity $v$ (scalar float number), and steering angle $\omega$ (scalar float number). Each dataset $\mathcal{D}$ is therefore 
$
\mathcal{D} = \{ i_t, w_t, g_t, d_t, s_t, v_t, \omega_t \}_{t=1}^N
$, where $N$ indicates the total number of data frames. 

\begin{figure}
  \centering
  \includegraphics[width=1\columnwidth]{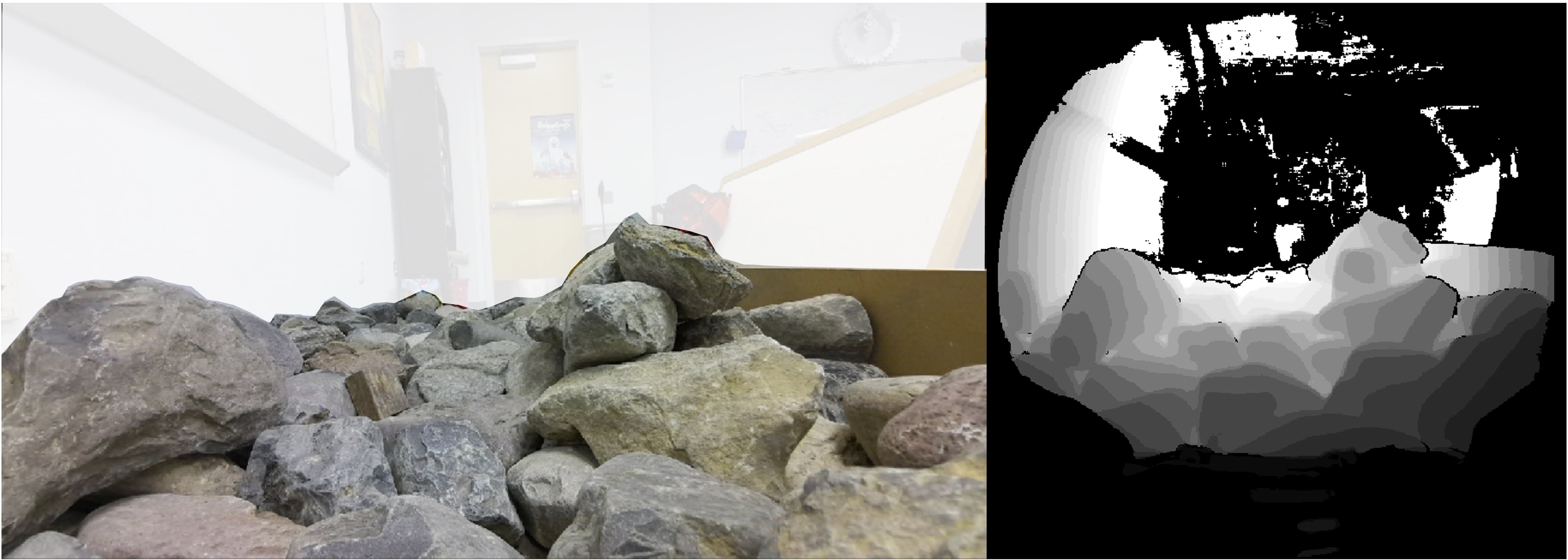}
  \caption{Example RGB (Left) and Depth (Right) Images in the Verti-Wheelers Dataset.}
  \label{fig::dataset}
\end{figure}

The initial release of the V6W and V4W datasets include 50 and 64 teleoperated trials of 46667 and 70143 data frames of the 6-wheeler and 4-wheeler crawling over different vertically challenging rock/boulder courses respectively. To assure the human demonstrator has only access to the same perception as the robot, teleoperation is conducted in a first-person-view from the onboard camera, rather than a third-person-view. Fig.~\ref{fig::dataset} shows an example data frame of the RGB and depth images $i$ of the V4W dataset. 

With a more mechanically capable chassis and two more wheels, the demonstrator can demonstrate crawling behaviors on V6W at ease, i.e., mostly driving forward, slowing down when approaching an elevated terrain patch, and only using the steering to circumvent very difficult obstacles or ditches in front of V6W. On the other hand, the demonstration of V4W takes much more effort, and the demonstrator needs to carefully control both the linear velocity and steering angle at fine resolution to negotiate through a variety of vertically challenging terrain. 

\section{ALGORITHMS}
\label{sec::algorithms}

Requiring no or little specialized hardware in addition to conventional wheeled platforms, we develop three algorithms to drive our robots through vertically challenging terrain: an open-loop, a classical rule-based, and an end-to-end learning-based controller. 

\subsection{Open-Loop Controller}
As a baseline, we implement an open-loop controller that drives the robots toward vertically challenging terrain previously deemed as non-traversal obstacles, simply treating them as free spaces. Our open-loop controller locks the differentials and uses the low gear all the time. We set a constant linear velocity to drive the robot forward. No onboard perception is used for the open-loop controller. 

\subsection{Classical Rule-Based Controller}
\begin{figure}
  \centering
  \includegraphics[width=\columnwidth]{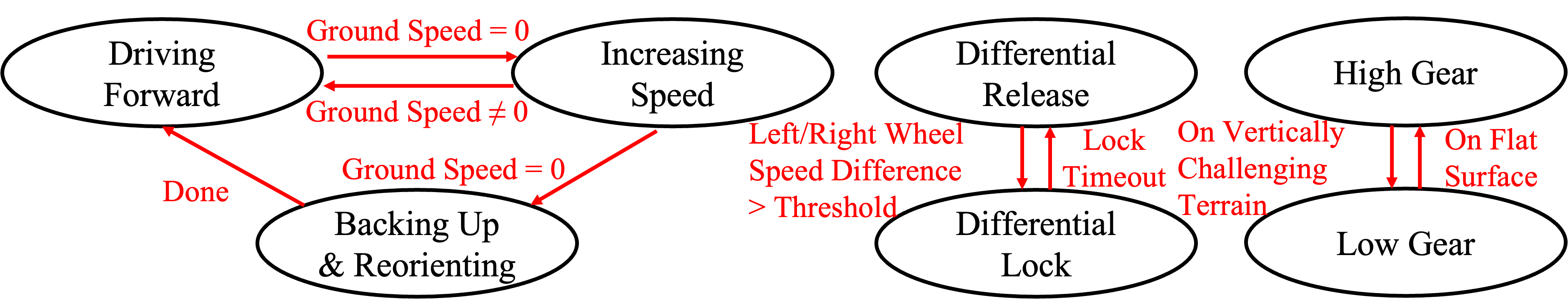}
  \caption{Finite State Machine of the Rule-Based Controller}
  \label{fig::fsm}
\end{figure}

We design a classical rule-based controller based on our heuristics on off-road driving: we lock the corresponding differential when we sense wheel slippage; we use the low gear when ascending steep slopes; when getting stuck on rugged terrain, we first increase the wheel speed and attempt to move the robot forward beyond the stuck point; if unsuccessful, we then back up the robot to get unstuck, and subsequently try a slightly different route. With the hardware described in Sec. \ref{sec::platforms}, our robots are able to perceive all aforementioned information with the onboard sensors and initiate corresponding actions. The finte state machine of the rule-based controller is shown in Fig.~\ref{fig::fsm}. 

\subsection{End-to-End Learning-Based Controller}
We also develop an end-to-end learning-based controller to enable data-driven mobility. We aim at learning a motion policy that maps from the robots' onboard perception to raw motor commands to drive the robots over vertically challenging terrain. Utilizing the datasets we collect, we adopt an imitation learning approach, i.e., Behavior Cloning~\cite{bojarski2016end}, to regress from perceived vehicle state information to demonstrated actions.

\begin{table*}[ht]
\caption{Number of Successful Trials (Out of 10) and Mean Traversal Time (of Successful Trials in Seconds) with Variance}
\centering
\begin{adjustbox}{width=1\textwidth}
\begin{tabular}{ccccccccc}
\toprule
                   & \multicolumn{4}{c}{\footnotesize V6W}                        & \multicolumn{4}{c}{\footnotesize V4W}                        \\
\cmidrule(rl){2-5} \cmidrule(rl){6-9}
                   & \footnotesize OL & \footnotesize RB & \footnotesize BC6 & \footnotesize BC4 & \footnotesize OL & \footnotesize RB & \footnotesize BC6 & \footnotesize BC4 \\
\midrule
\footnotesize Easy      & 5  ($20.7\pm1.7$) & 8  ($19.2\pm3.9$) & 9  ($13.8\pm8.2$) & \textbf{10}  ($11.6\pm1.9$)  & 6  ($17.7\pm3.8$) & 6  ($13.4\pm2.5$) & 7  ($17.2\pm6.7$)  & \textbf{9}  ($14.1\pm7.7$) \\
\footnotesize Medium    & 6  ($15.4\pm0.9$) & 9  ($14.8\pm2.2$) & 9  ($14.6\pm11.2$) & \textbf{10}  ($13.6\pm2.3$)  & 4  ($15.6\pm14.2$) & 6  ($12.9\pm1.8$) & 3  ($19.2\pm10.6$) & \textbf{8}  ($13.7\pm1.6$) \\
\footnotesize Difficult & 3  ($24.1\pm2.6$) & 6  ($14.3\pm1.9$) & 6  ($15.7\pm18.5$) & \textbf{9}  ($14.9\pm2.9$) &   3  ($19.7\pm29.4$) & 5  ($16.8\pm20.5$) & 3  ($23.3\pm43.4$) & \textbf{7}  ($14.9\pm8.2$) \\
\bottomrule
\end{tabular}
\end{adjustbox}
\label{tab::results}
\end{table*}

Our motion policy $\pi_\theta: \mathcal{X} \rightarrow \mathcal{A}$, is a mapping from vehicle state space $\mathcal{X}$ to vehicle action space $\mathcal{A}$, parameterized by a set of parameters $\theta$. To instantiate a vehicle state $x \in \mathcal{X}$, we use all (or a subset of) onboard perception, including RGB and depth image $i$, wheel speed $w$, and ground speed $g$, while the vehicle action $a \in \mathcal{A}$ includes differential release/lock $d$, low/high gear switch $s$, and most importantly, linear velocity $v$ and steering angle $\omega$:  
\begin{equation}
x = (i, w, g), ~
a = (d, s, v, \omega).
\label{eqn::state}
\end{equation}
To train $\pi_\theta$, we minimize an end-to-end behavior cloning loss to seek for the optimal parameter set $\theta^*$:
\begin{equation}
    \begin{split}
    \theta^* &= \argmin_{\theta} \sum_{(x, a) \in \mathcal{D}} ||a - \pi_{\theta}(x)||_H,
    \end{split}
    \label{eqn::bc}
\end{equation}
where $||v||_H = v^THv$ is the induced norm by a diagonal matrix $H$ with positive real entries to weigh each dimension of the action. We instantiate $\pi_\theta$ as a neural network and use backpropagation to find the appropriate set of parameters $\theta^*$.

\section{EXPERIMENTS}
\label{sec::experiments}


All three algorithms in Sec. \ref{sec::algorithms} are implemented on both V6W and V4W in Sec. \ref{sec::platforms}, with the end-to-end learning-based controller learned from the collected datasets. 

\subsection{Implementations}
 Considering the extremely rugged testbed (Fig.~\ref{fig::testbed}), all three methods lock both differentials and use low gear. 
The open-loop controller (OL) uses a constant linear velocity of 0.5m/s. When the rule-based (RB) controller gets stuck for 2 seconds, the wheel speed is gradually increased from 0.5m/s to 0.75m/s. If unsuccessful for 3 seconds, the robots back up for 2 seconds at 0.5m/s, 
and then continue forward at 0.75m/s steering 18\textdegree \space left for 2 seconds and right for 2 seconds.

For the end-to-end Behavior Cloning (BC) controller, we use both datasets on both vehicles. Specifically, in addition to using their own corresponding dataset, we also cross-deploy the learned model on the other platform, i.e., learned model from V6W data deployed on V4W and vice versa. Such a cross-deployment aims at revealing the relationship between different mechanical capabilities and qualities of the training data. Similar to OL and RB, BC maintains locked differentials and low gear. Our BC implementation takes vehicle state $x=(i)$ (depth image only) and action $a=(v, \omega)$ (Eqn. \ref{eqn::state}) and seeks optimal parameter $\theta^*$ (Eqn. \ref{eqn::bc}) in a neural network. BC uses a RESNET18 to process the $224\times224$ depth images (down sampled from $512\times512$) before feeding the 512-dimensional embedding into four fully connected layers with 256, 128, 64, and 2 neurons with ReLU activation except the last layer. 

\subsection{Results}
Three different test courses are built by reconfiguring the rocks/boulders on the testbed (Fig.~\ref{fig::testbed}), whose difficulty levels range from easy, medium, to difficult. For the difficult level, we add wooden blocks to the rock/boulder course, which do not exist in the datasets, to test BC's generalizability. For each test course, we run four different approaches, i.e., OL, RB, BC, and BC with cross-deployment (we denote the model trained with the V6W and V4W dataset as BC6 and BC4 respectively), each ten trials, running from both directions of the course. We report both number of successful trials (out of 10 attempts) and mean traversal time (for the successful trials in seconds) with variance of all 240 experiment trials in Tab. \ref{tab::results}. A failure trial can either be the vehicle getting stuck or tipping over on the test course. For all four approaches, we start the vehicles at the same starting location and orientation facing the test course. 

For a certain vehicle on a certain difficulty level, in general BC achieves higher success rate than both OL and RB, with OL most frequently getting stuck or tipping over. Among all successful trials, BC mostly achieves the shortest traversal time, but not always, because BC learns to slow down to smoothly go through rugged terrain while OL and RB may drive aggressively. With increasing difficulty level, success rate decreases and traversal time increases, although V6W's performance is not very sensitive to the Easy and Medium levels, considering its superior mechanical capability. 

One very interesting finding is the cross-deployment results. While V4W trained with its own dataset (BC4) outperforms V4W cross-trained with V6W's dataset (BC6), V6W's performance is opposite: V6W performs better when cross-trained with V4W's dataset (BC4) compared to when being trained with its own dataset (BC6). Such a discrepancy is caused by the different demonstrations in the V6W and V4W datasets, which are further caused by the difference in the two vehicles' mechanical capabilities: as mentioned above, the demonstrator can drive the more mechanically capable V6W through different test courses at ease without the need to constantly adjust the speed and steering, while the limited mechanical capability of V4W requires fine-grained control, causing the quality of the BC4's dataset to exceed that of BC6's. Therefore, even V6W, a different vehicle, can achieve improved mobility with BC4, compared to BC6 with its own dataset. On the other hand, learning from the less careful driving demonstrations with BC6 jeopardizes the learned mobility of V4W. 

\subsection{Outdoor Mobility Demonstration}
\begin{figure}
  \centering
  \includegraphics[height=0.39\columnwidth]{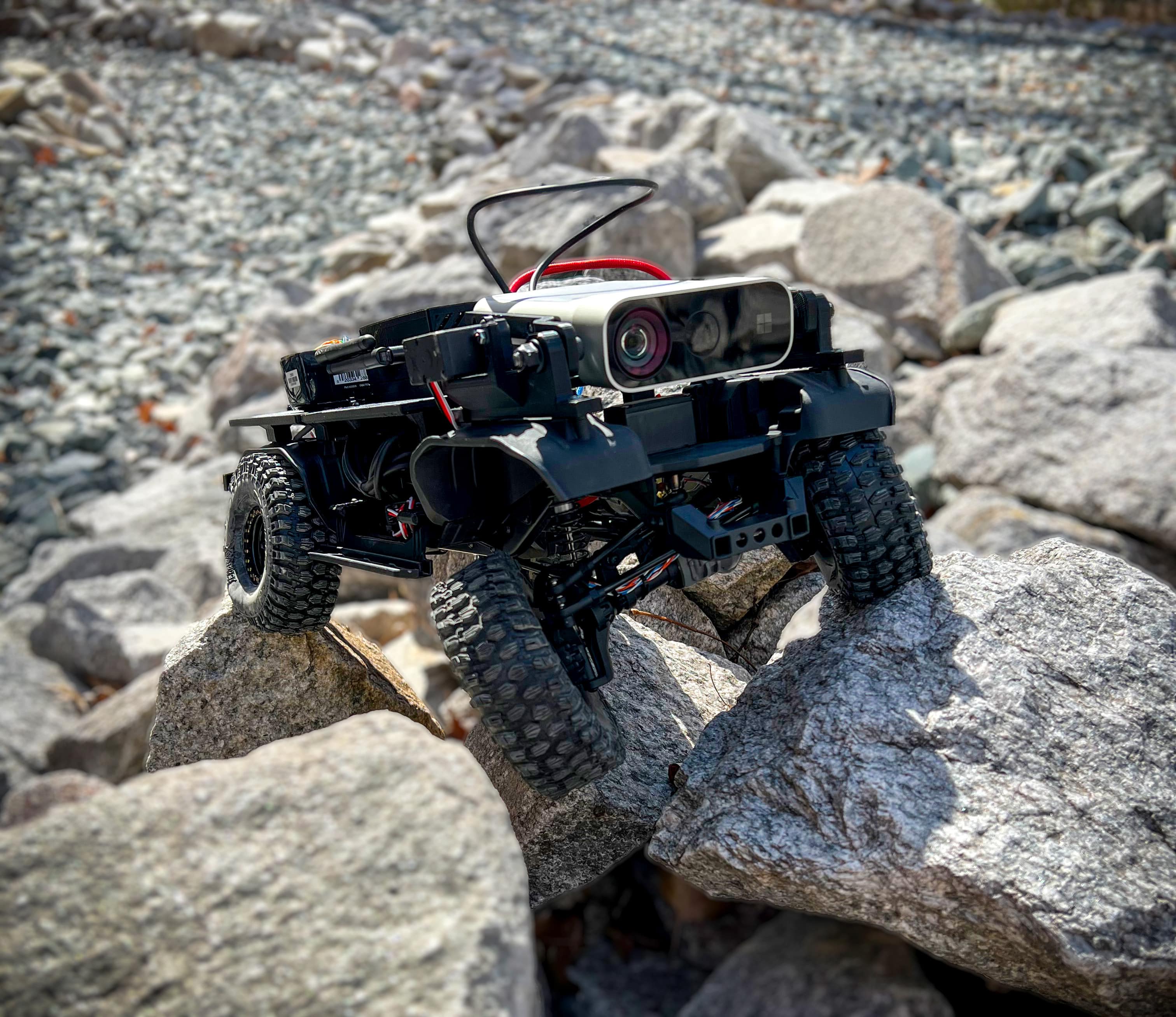}
  \includegraphics[height=0.39\columnwidth]{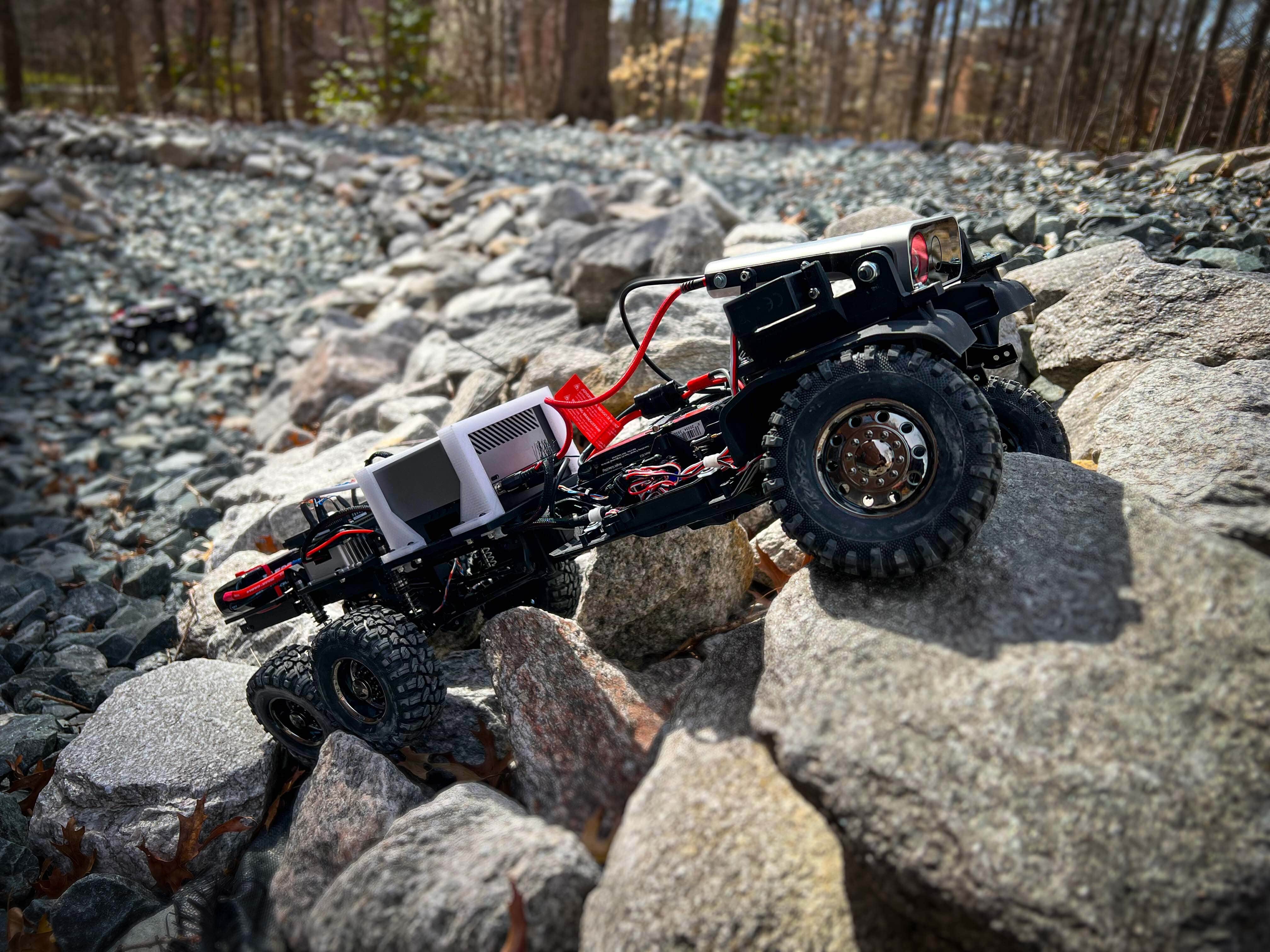}
  \caption{Outdoor Demonstration: V4W's right front wheel suspends in the air (left) and V6W climbs a steeper rock pile than the indoor testbed. }
  \label{fig::outdoor}
\end{figure}

We also deploy both Verti-Wheelers in outdoor vertically challenging environments beyond the controlled indoor testbed (Fig.~\ref{fig::outdoor}). The outdoor environments contain different rock sizes, including both smaller pebbles and larger boulders compared to the indoor testbed. Furthermore, the outdoor environments are more jagged, including larger ditches between rocks that may easily get the vehicles stuck (Fig.~\ref{fig::outdoor} left) and steeper slopes that require stable traction between all wheels and the terrain (Fig.~\ref{fig::outdoor} right). While OL and RB can drive smoothly on small pebbles, they often get stuck on larger and more jagged boulders. BC is able to generalize well to rocks of similar sizes as in the indoor testbed, but may occasionally get stuck on large boulders, wide ditches, and steep slopes as well, showing future research is still required to improve wheeled mobility on vertically challenging terrain in the real world. 
\section{CONCLUSIONS AND FUTURE WORK}
\label{sec::conclusions}
This paper presents two vehicle platforms, demonstration datasets, and three algorithms to unlock the previously unrealized potential of wheeled mobility on vertically challenging terrain. The two platforms, with minimal hardware modification to conventional wheeled robots, achieve seven desiderata to tackle such challenging environments. We open-source our hardware design, software implementation, and demonstration datasets to facilitate future research. The experiment results of the three algorithms confirm our hypothesis that conventional wheeled robots have the mechanical capability of navigating many vertically challenging terrain, which are normally considered as non-traversable obstacles, especially with the help of data-driven approaches. 

This paper opens up a new direction of research to achieve superior robot mobility beyond their original design with only limited mechanical capability. Despite the fact that the simple Behavior Cloning algorithm works well on many vertically challenging terrain, it still fails in many more challenging scenarios, such as in more difficult test courses than the Difficult level in Tab.~\ref{tab::results} or in the natural outdoor environments (Fig.~\ref{fig::outdoor}). However, these environments can be conquered by human teleoperation and are therefore within the vehicles' mechanical limit. One promising future research direction is to explicitly model terrain traversability instead of implicitly learning it in a motion policy. It is also possible to design or learn a vehicle dynamics model on vertically challenging terrain. Another interesting direction is to combine autonomous crawling and goal-oriented navigation with obstacle avoidance, i.e., real obstacles that are absolutely beyond the robots' mechanical capability. Last but not least, considering the interesting findings in the BC cross-deployment experiments, Verti-Wheelers provide ideal physical robot platforms for research in transfer learning for cross-platform mobility, learning with embodiment mismatch, and meta learning for general vehicle mobility and maneuverability. 

\section*{ACKNOWLEDGEMENTS}
This work has taken place in the RobotiXX Laboratory at George Mason University. RobotiXX research is supported by Army Research Office (ARO, W911NF2220242, W911NF2320004, W911NF2420027), US Air Forces Central (AFCENT), Google DeepMind (GDM), Clearpath Robotics, and Raytheon Technologies (RTX). 

\bibliographystyle{IEEEtran}
\bibliography{IEEEabrv,references}
\end{document}